\pdfoutput=1

\documentclass[11pt]{article}

\usepackage{ACL2023}

\usepackage{times}
\usepackage{latexsym}
\usepackage[T1]{fontenc}
\usepackage[utf8]{inputenc}
\usepackage{microtype}
\usepackage{inconsolata}

\usepackage{array}

\usepackage[TABBOTCAP]{subfigure}
\usepackage[shortlabels]{enumitem}
\usepackage{placeins}
\usepackage{bbold}
\usepackage{tikz-dependency}
\usepackage{algorithm}
\usepackage{algpseudocode}
\usepackage{multirow}
\usepackage{booktabs}

\usepackage{color}
\usepackage{helvet}
\usepackage{textcomp}
\usepackage{booktabs}

\usepackage{color}
\usepackage{bbm}
\usepackage{graphicx}
\graphicspath{ {images/} }
\usepackage{amsmath}
\usepackage{amsfonts}

\usepackage{hyperref}
\usepackage{url}
\usepackage{tabularx}

\newcommand{\Red}[1]{\textcolor[rgb]{1.00,0.00,0.00}{#1}}

\newcommand{\Green}[1]{\textcolor[rgb]{0.00,0.80,0.00}{#1}}

\usepackage{xcolor}
\usepackage{color-edits}
\addauthor{hm}{red}
\addauthor{sw}{orange}
\addauthor{xr}{cyan}

\title{Fact-and-Reflection (FaR) Improves Confidence Calibration \\ of Large Language Models}
\author{
 \textbf{Xinran Zhao$^{1,2,}$}\thanks{\quad Work done during an internship at Tencent AI Lab, Bellevue. Corresponding contact email addresses: \{xinranz3,sherryw\}@andrew.cmu.edu, \{hongmingzhang, xiaomanpan ,wenlinyao,dyu,jianshuchen\}@global.tencent.com. Our code is publicly available at: \url{https://github.com/colinzhaoust/fact-and-reflection}.}\quad
 \textbf{Hongming Zhang$^1$}\quad
 \textbf{Xiaoman Pan$^1$}\quad
 \textbf{Wenlin Yao$^1$}\quad \\
 \textbf{Dong Yu$^1$}\quad
  \textbf{Tongshuang Wu$^2$}\quad
 \textbf{Jianshu Chen$^1$}\\
 $^1$Tencent AI Lab, Bellevue, $^2$Carnegie Mellon University
}

\date{}

\begin{document}
\maketitle
\begin{abstract}
For a LLM to be trustworthy,
its confidence level should be \emph{well calibrated} with its actual performance. 
While it is now common sense that LLM performances are greatly impacted by prompts, the confidence calibration in prompting LLMs has yet to be thoroughly explored.
In this paper, we explore how different prompting strategies influence LLM confidence calibration and how it could be improved. 
We conduct extensive experiments on six prompting methods in the question-answering context and we observe that, while these methods help improve the expected LLM calibration, they also trigger LLMs to be \textit{over-confident} when responding to some instances.
Inspired by human cognition, we propose Fact-and-Reflection (FaR) prompting, which improves the LLM calibration in two steps. First, FaR elicits the known ``facts'' that are relevant to the input prompt from the LLM. And then it asks the model to ``reflect'' over them to generate the final answer.
Experiments show that FaR prompting achieves significantly better calibration; it lowers the Expected Calibration Error by 23.5\% on our multi-purpose QA tasks. 
Notably, FaR prompting even elicits the capability of verbally \textit{expressing concerns} in less confident scenarios, 
which helps trigger retrieval augmentation for solving these harder instances.

\end{abstract}

\section{Introduction}
\begin{figure}[!t]
    \centering
    \includegraphics[clip,trim={0cm 19.5cm 43.5cm 0cm},width=0.9\linewidth]{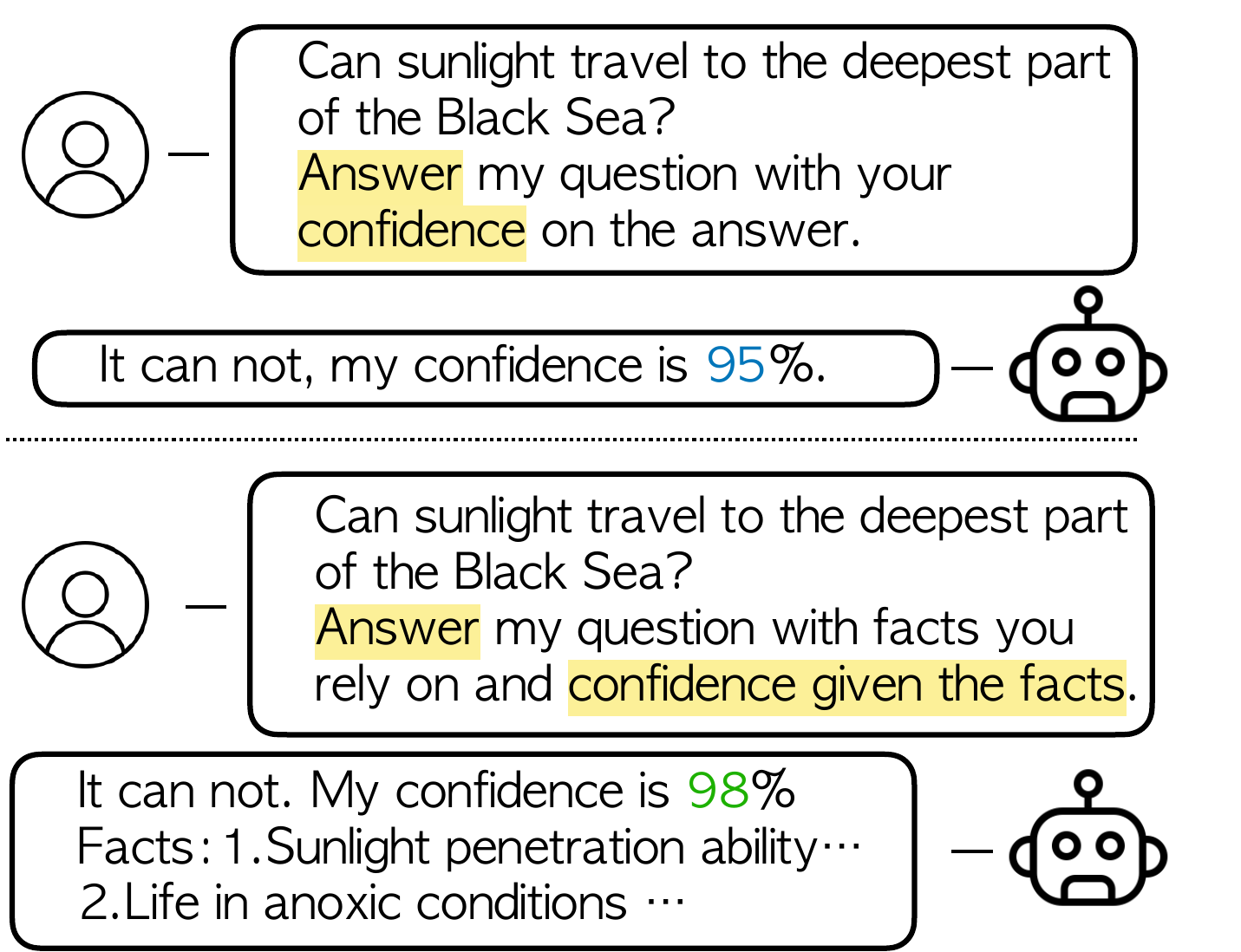}
    
    \caption{An example of how different prompting methods affect confidence extraction: model verbalized confidence of GPT-3.5 increases when asking the model to elicit the facts it relies on to answer the question.}
    \label{fig:motivation}
\end{figure}

With the emergence of Large Language Models (LLMs)~\citep{chowdhery2022palm, thoppilan2022lamda, openai2022chatgpt, openai2023gpt4, touvron2023llama, anil2023palm}, various prompting strategies have been proposed for \emph{improving the LLM performance}.
It is now common sense that well-designed prompts can help elicit desirable capabilities from LLMs that are important for various tasks. For example, Chain-of-Thought and its variants have been widely used for unlocking LLM reasoning capabilities~\citep{liu-etal-2022-generated, weichain, wang2023selfconsistency, press2023measuring, yao2023tree}.


However, what is less studied is the fact that prompting also influences the model confidence in their responses.
For example, as shown in Figure~\ref{fig:motivation}, the LLM's verbalized confidence~\citep{lin2022teaching} would shift when it is asked to simply first seek supporting facts. 
Effective \emph{confidence calibration} of LLMs --- output confidence scores matching model performance --- is crucial because calibrated confidence ensures the model reliability and helps guide practical use cases, e.g., identifying potential hallucinations
~\citep{kadavath2022language,varshney2023stitch}, applying additional fact-checking 
~\cite{chen-etal-2021-improving}, etc.
Beyond inference time usage, it can also help guide the training process, e.g., improving instruction tuning~\cite{chung2022scaling}.
Therefore, one central question we want to answer is: \emph{how do different prompting methods influence the confidence calibration?}




To do so, we first start by assessing six different prompting strategies, on Question-Answering (QA) datasets that rely on reasoning~\citep{geva2021strategyqa} and knowledge~\citep{bordes2014question}.
In our assessment, we employ \emph{confidence calibration}, specifically Expected Calibration Error (ECE) \citep{guo2017calibration} and Macro Calibration Error (MacroCE) \citep{si-etal-2022-examining}, to measure the discrepancy between a model's performance and its confidence levels.
They are widely recognized metrics for evaluating the quality of confidence scores in models, as discussed in prior studies~\cite{desai-durrett-2020-calibration, si2022prompting}.
We find that different prompting methods generally suffer from \emph{over-confidence}, and exhibit poor calibration at the instance level.



\begin{figure}[!t]
    \centering
    \includegraphics[clip,trim={0cm 18.4cm 41cm 0cm},width=0.9\linewidth]{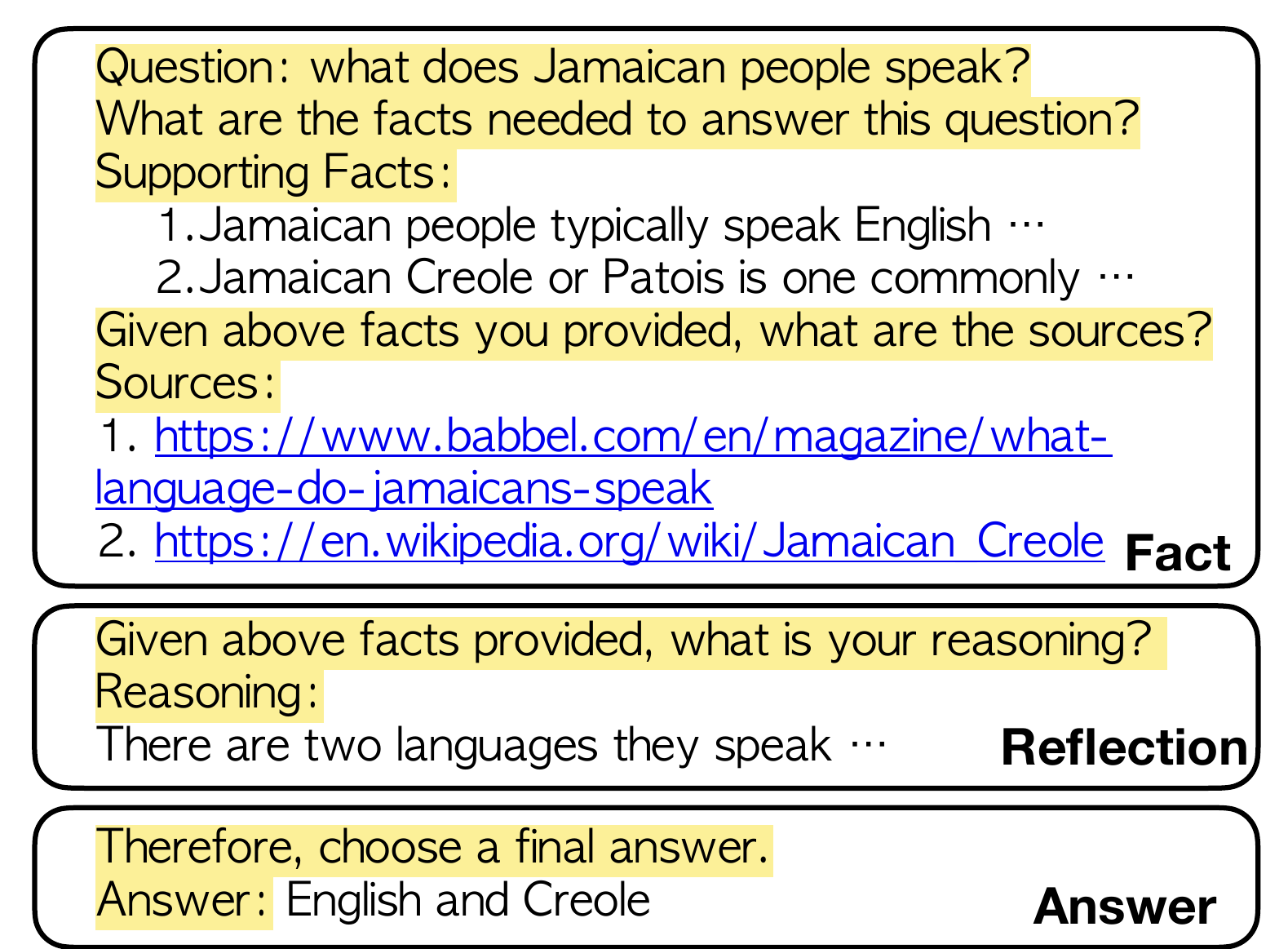}
    \caption{An example of the proposed FaR prompting, which consists of three steps: fact, reflection, and answer. Before answering the question and extracting the confidence scores, FaR explicitly decomposes the entire process into fact elicitation (with corresponding sources) and reflectively reasoning over the facts. The answers will then be utilized in the prompt to generate the final answer. Highlighted text denotes the input prompts provided to the model.}
    \label{fig:fsr}
\end{figure}

Psychological and cognitive research~\cite{BLOCK1991188, GEORGE2000195} indicates that human's over-confidence can be mitigated by disentangling the processes of fact acquisition and reasoning. Instead of reasoning while stating the facts, explicitly recalling all relevant facts before deliberation can avoid early \emph{anchoring} the reasoning process onto the first upcoming fact in the context. 
Accordingly, we propose our \textbf{F}act-\textbf{a}nd-\textbf{R}eflection (FaR) prompting to improve model confidence calibration --- see Figure~\ref{fig:fsr} for an example. Specifically, it consists of three steps: first, prompting the model for potential known facts and their sources, second, eliciting reflective reasoning to connect all recalled knowledge, and finally, generating the answer.

Our experiments on aforementioned datasets demonstrate that FaR prompting significantly reduces the confidence calibration error across various calibration measures (23.5\% and 13.9\% under ECE and MacroCE, respectively).

Further analysis reveals that the improvement comes from that FaR prompting intrigues the model to generate cautious answers that express concerns, such as adding a comment like ``there is no sufficient evidence'' after the answer. We observe that \textit{expressing concerns} co-occurs with an average of 13.2\% reduction in confidence relative to the situations without \textit{expressing concerns}. This phenomenon can suggest a potential mechanism that helps detect hard instances that are not answerable with LLM's internal knowledge and may benefit from retrieval augmented generation. 

In summary, our main contributions are:
\begin{enumerate}[noitemsep,labelwidth=*,leftmargin=1.8em,align=left]
    \item We study how different prompting methods influence confidence calibration and find that many methods, though generally being helpful, suffer from the over-confidence issue.
    \item We propose FaR prompting that improves model confidence calibration across various common metrics. 
    \item We show that FaR prompting mitigates \textit{over-confidence}. Moreover, FaR prompting elicits the model to express concerns when answering questions they are uncertain of.
\end{enumerate}

\section{How do Prompting Strategies Influence Confidence Calibration in LLMs?}
\label{sec:impact_of_prompting_methods}



To examine the influence of different prompting strategies on the confidence calibration of LLMs, we follow \citet{tian2023just} and conduct our experiments on QA. 
Specifically, we use QA datasets that mainly examine either reasoning~\citep[StrategyQA, ][]{geva2021strategyqa} or knowledge~\citep[Web Questions. ][]{bordes2014question} where the confidence calibration is important due to the reliability requirement. An example question would be, ``Did Aristotle use a laptop?''. The answers are either in the format of Yes/No or a set of short phrases, e.g., ``English and Creole''. 
We report the macro-average scores for two kinds of QA data.
Following \citet{wei2022emergent}, we conduct all of our experiments using OpenAI's GPT-3~(\texttt{text-davinci-003},~\citealp{ouyang2022training})\footnote{At the time of writing, this model was the only one that both has sufficient reasoning capability for solving StrategyQA and provides access to the logits (necessary for confidence calculation). Experiments with other models are in Section~\ref{sec:generalize_backbone}.}. We set the default max output length to 120 and the temperature to 1.2.

\subsection{Evaluation Metrics}

We follow the conventional approach ~\citep{guo2017calibration,desai-durrett-2020-calibration,si-etal-2022-examining}, and measure the confidence score quality with \emph{confidence calibration}: the error between \emph{model performance} and \emph{confidence scores} across instances. Lower errors indicate better calibration and thus, higher accuracy in confidence scores.

We measure \textbf{model performance} with the exact match criterion, where an answer is deemed correct if it matches a candidate label after normalization.


Then, following~\citet{si-etal-2022-examining}, we use both Expected Calibration Error~\citep[ECE, ][]{guo2017calibration} and Macro-average Calibration Error~\citep[MacroCE, ][]{si-etal-2022-examining} to evaluate the quality of \textbf{confidence calibration}.
Both metrics are based on the difference between the confidence scores and the correctness of model predictions. 
\begin{itemize}[noitemsep, nosep,labelwidth=*,leftmargin=*,align=left]

\item \texttt{ECE} uses a bucketing approach that measures the \emph{overall calibration}.
It assigns examples with similar confidence to the same buckets.
Given the input $x$, ground truth $y$, prediction $\tilde{y}$, and $B_m$ denoting the $m$-th bucket for ($x$,$y$,$\tilde{y}$), for $N$ model predictions bucketed into $M$ buckets: $$\text{ECE} = \frac{1}{N}\sum_{m=1}^M |B_m| \cdot |\text{Acc}(B_m) - \text{Conf}(B_m)|,$$ where $\text{Acc}(B_m)$ and $\text{Conf}(B_m)$ denote the accuracy and averaged confidence for the samples in $B_m$, and $|B_m|$ denotes the cardinality of $B_m$.
Such definition triggers \emph{bucket-canceling effect}, i.e., the over- and under- confident instances within the same bucket may cancel with each other and hence not contribute to the overall error.
As a result, ECE provides \textbf{a stable overall measure} of how well the confidence matches the expected accuracy --- 
A single outlier (e.g., extremely high confidence for a wrong prediction) will not affect the global ECE too much.

\item \texttt{MacroCE} is the (macro) average of the following two instance-level calibration errors (ICE), which measures the ICE for the $N_p$ correctly and $N_n$ incorrectly predicted samples, respectively:
$$\text{ICE}_\text{pos} = \frac{1}{N_p} \sum_{i=1}^{N_p}(1-\text{Conf}(x_i,\tilde{y}_i)), \forall_i y_i= \tilde{y}_i,$$
$$\text{ICE}_\text{neg} = \frac{1}{N_n} \sum_{i=1}^{N_n}(\text{Conf}(x_i,\tilde{y}_i) - 0), \forall_i y_i \neq \tilde{y}_i.$$

\end{itemize}

MacroCE does not have the bucket-canceling effects, provides more granularity, and is more reflective of \textbf{instance-level confidence calibration}.
For example, consider a set of two predictions $\{1,0\}$ with labels $\{0,0\}$. When the corresponding confidence scores are $\{1,0\}$, assume they are in the same bucket, the output of ECE will be 0, as the difference between the average confidence scores and accuracy. Therefore, the high error extreme instances are not captured by ECE (e.g., wrong, but $\text{conf}=1$). In contrast, each instance contributes to MacroCE (equals 1 in this case), which reveals more instance-level effects than ECE. 

\subsection{Prompting Methods}
\label{sec:baselines}

We examine how different prompting methods influence confidence calibration. Besides \emph{Standard} prompting, we categorize the strategies into two kinds: Step Decomposition and Multi-Candidate Selection (examples in Figure~\ref{fig:prompt_examples} in the appendix).%

\textbf{Step Decomposition} prompts guide the model to output multiple steps of thoughts or knowledge that may facilitate answering the question in the final step. We consider three common ones:

\begin{enumerate}[noitemsep, nosep,labelwidth=*,leftmargin=*,align=left, series=prompts]

\item \texttt{Knowledge prompting}~\citep{liu-etal-2022-generated}: We insert a prompt ``Generate some knowledge about the question:'' after the original question, and have the model generate the final answer based on the generated knowledge. We also explore a minor variant: \textit{Knowledge+Explain}, which denotes changing the final prompt from ``Answer:'' to ``Explain and Answer:''. 

\item \texttt{Chain-of-Thought} prompting ~\citep[CoT,][]{weichain}: We have the model generate chained reasoning before the final answer. We follow~\cite{kojima2023large} to conduct zero-shot CoT by adding a prompt ``Let's think step by step:'' after the original question.

\item \texttt{Self-ask}~\citep{press2023measuring}: we have the model decompose the original question into multiple sub-questions, generate the intermediate answers, and combine them in the prompt to get the final answer. In detail, we first ask the model ``Are follow-up questions needed?''. If not, we directly get the final answer; Otherwise, we ask the model to generate the follow-up questions and the corresponding intermediate answers. All these intermediate steps are included in the final prompt (``Question:<question>; Intermediate Questions and Answers: <generated question-answer pairs> Answer:''). \textit{Self-Ask (aggregate)} denotes the variant that asks the model to answer all the intermediate questions together in a one-step generation. 
\end{enumerate}

On the other hand, \textbf{Multi-candidate Selection} methods acquire the final answer through querying the model for multiple rounds and selecting a candidate with pseudo discussion or majority voting. We examine the following two (we extract their confidence scores after the final answer):

\begin{enumerate}[resume*=prompts]

\item \texttt{Self-consistency}~\citep{wang2023selfconsistency}: We have the model generate the answers multiple times with the standard prompt, and select the final answer through a majority vote. We follow the original work to sample 10 different outputs with a temperature 0.7.

\item \texttt{Tree-of-Thought} prompting~\citep[ToT,][]{yao2023tree}: We follow the naming protocol by \citet{tree-of-thought-prompting}, and have the model to perform state-aware generation and search, by mimicking multiple experts discussing the input questions in pseudo conversations. We denote our method as pseudo-ToT since the search steps are not conducted with a separate module.
\end{enumerate}








\begin{figure}[t]
    \centering
    \includegraphics[clip,trim={0cm 19cm 40cm 0cm},width=0.9\linewidth]{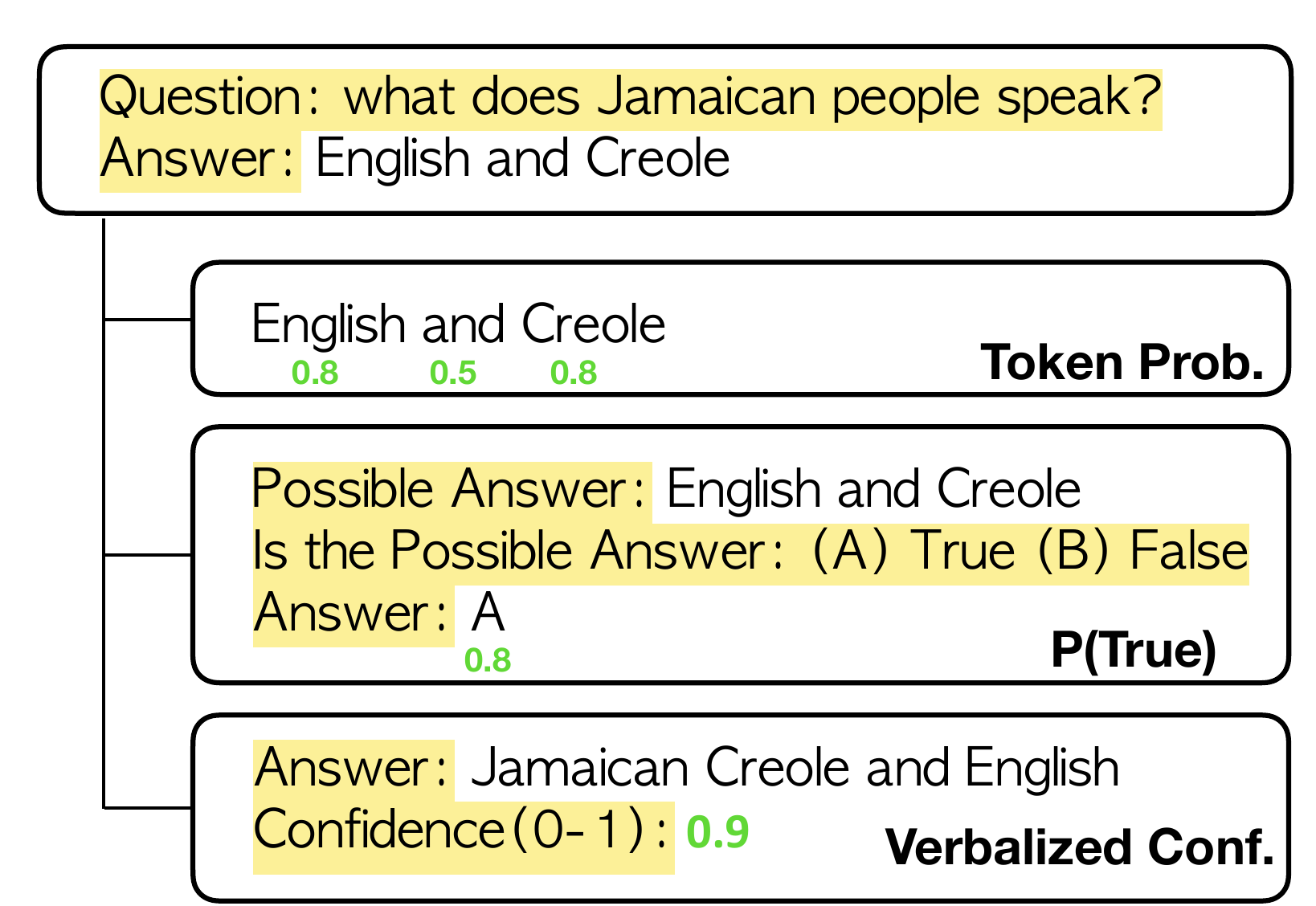}

    \caption{Different methods for computing confidence scores. The numbers in green denote the token probability of the output sequences. Highlighted texts denote input prompts provided to the model. Note that Both P(True) and Verbalized Confidence can be considered as suffix prompting methods that obtain the confidence scores after the model outputs its answer.}
    \label{fig:conf_extract}
\end{figure}

\subsection{Confidence Extraction Methods}
\label{sec:conf_extract}

To obtain \textbf{confidence scores}, we examine three widely used methods in literatures (Figure~\ref{fig:conf_extract}):

\begin{itemize}[noitemsep, nosep,labelwidth=*,leftmargin=*,align=left]
\item \texttt{Token Prob.}: Computed by averaging the top-1 log probability of each token over the entire sequence, and then applying the exponential. This is the reciprocal of the perplexity of the generated sequence with greedy decoding.

\item \texttt{P(True)}~\cite{kadavath2022language}: After prompting the model to generate the answer with ``\emph{Possible answer: <model\_answer>}'', it appends a suffix prompt asking ``\emph{Is the possible answer: (A) True (B) False}'', and then extracts the probability of answering ``\emph{A}'' as the confidence score.

\item \texttt{Verbalized Confidence} \citep{lin2022teaching}: After the model generates the answer, it further prompts the model to generate the confidence score by using suffix ``\emph{Confidence (0-1):}''\footnote{Slightly different from \citet{lin2022teaching}, we further add the hint for the range of confidence score ``(0-1)''.}.
\end{itemize}


\begin{table}[t]
\small
    \centering
    \begin{tabular}{l|c|c}
    \toprule
        Conf. Extraction & ECE $\downarrow$  & MacroCE $\downarrow$ \\
         \midrule
         Token Prob. & 27.3 & 80.8 \\
         P(True) & 41.3 & 70.3 \\
         Verbalized & 42.7 & 106.3\\
         \bottomrule
    \end{tabular}
    \caption{Expected Calibration Error (ECE) and Macro-average Calibration Error (MacroCE) of different confidence extraction methods in Table~\ref{tab:prompt_style}. Results are averaged over different prompting methods. Down arrows indicate the lower is the better.}
    \label{tab:conf_extraction_methods}
     \vspace{-0.2in}
\end{table}

To validate how applicable these methods are in our scenario, we first compare different confidence extraction methods to determine the ones to be used in the later experiments. We use the ECE and MacroCE metrics averaged over different prompting methods to measure the overall and instance-level performance (the lower, the better). The average is calculated over the 8 baseline methods in Table~\ref{tab:prompt_style}.


Table~\ref{tab:conf_extraction_methods} shows that overall, Token Prob. achieves the best ECE scores. On the other hand, P(True) achieves the best MacroCE scores. The reason can be that, while P(True) consistently measures the probability of an option for a multiple-choice question, Token Prob. can be unstable when the generated answer is long and includes auxiliary words (``English and Creole'' vs. ``Jamaican people speak English and Creole'' in Figure~\ref{fig:conf_extract}). 
Verbalized Confidence shows the worst performance among the confidence extraction methods, indicating that further improvement is still required (e.g., tailor-made instructions shown in \citet{tian2023just}). Therefore, we exclude Verbalized Confidence extraction method in our following experiments, and for simplicity, we only report an aggregated overall performance of calibration by averaging the metrics obtained from using Token Prob. and P(True)\footnote{See Table \ref{tab:full_performance} in the appendix for the full results. Since the confidence scores are on the same scale (0-1) and a specific calibration metric (e.g., ECE) essentially measures the same kinds of error regardless of the confidence extraction methods, reporting the average measures the overall performance.}.

\begin{table}[t]
\small
    \centering
    \begin{tabular}{l|c|c}
    \toprule
        Prompting Method  & ECE $\downarrow$ & MacroCE $\downarrow$\\
        \midrule
        Standard  & 30.3 & \textbf{54.6} \\ %
        \midrule
        \multicolumn{3}{c}{ \textit{Step Decomposition}}\\
        \midrule
        Knowledge  & 33.0 & 73.9 \\
        Knowledge+Explain  & 27.1 & 64.5 \\
        CoT  & 29.6 & 62.3 \\
        Self-Ask  & 26.4 & 66.6 \\
        Self-Ask (aggregate)  & \textbf{26.0} & 66.0 \\
        \midrule
        \multicolumn{3}{c}{ \textit{Multi-Candidate Selection}}\\
        \midrule
        Self-Con.  & 34.7 & 67.6 \\
        Pseudo-ToT  & 33.0 & 73.5 \\

         \bottomrule
    \end{tabular}
    \caption{Expected Calibration Error (ECE) and Macro-average Calibration Error (MacroCE) of different prompting methods of different categories, as introduced in Section~\ref{sec:baselines}. The down arrow implies the lower, the better. The best-performing entry on each column is marked in \textbf{bold}.} 
    \label{tab:prompt_style} 
\end{table}

\subsection{Impact of Prompting Methods}
\label{sec:prompt_methods}
Table~\ref{tab:prompt_style} shows the impact of prompting methods on confidence calibration. We observe: 


\emph{Step Decomposition helps improve the model's overall expected confidence calibration} (Knowledge+Explain, CoT, Self-Ask), as reflected by ECE in Table~\ref{tab:prompt_style}. The reason can be that, with the generated chained thoughts or intermediate question-answer pairs in the context, the model-generated confidence is grounded onto the elicited thoughts.

\emph{Multi-Candidate Selection may cast a negative effect on the confidence calibration} (Self-Con., Pseudo-ToT). The reason can be that the candidate proposal stage adds to the randomness of the context. The final answers are sampled from and not necessarily aware of other candidates when generated, but the confidence extraction methods are conditioned on all candidates. Such mismatch can lead to lower confidence calibration and suggests that specialized confidence extraction methods should be developed to improve their performance.

Compared to the standard prompting, \emph{the advanced prompting methods (e.g., CoT) seem to degrade the instance-level calibration} i.e., they can simultaneously achieve higher MacroCE but lower ECEs.
Such mismatch suggests that instance-level extreme values have a negative impact on the confidence calibration. 
We then raise the question: what are the major causes of these high error extreme values: over-confidence (high confidence in wrong answers) or under-confidence (low confidence in correct answers)? 
We select the best-performing prompting methods to investigate further.

\paragraph{Over-Confidence Degrades Calibration}

\begin{figure}[!t]
    \centering %
    \includegraphics[clip,trim={0cm 1cm 0cm 0cm},width=\linewidth]{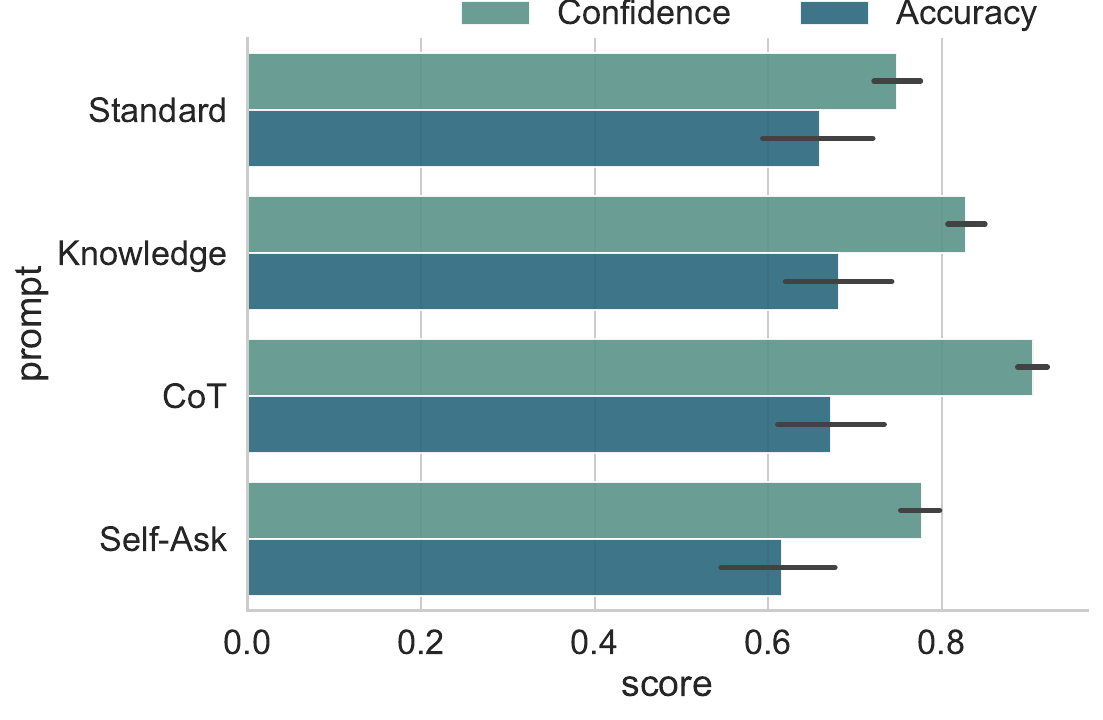}
    \caption{Accuracy versus confidence scores. Higher confidence scores relative to accuracy mean that the model is generally over-confident. In the ideal calibration case, there should be no gap between them.}
    \label{fig:over_under_confidence}
\end{figure}

We compare the averaged confidence scores (over all the samples) and accuracy in Figure~\ref{fig:over_under_confidence}. 
In the ideal case where we reach the perfect instance-level calibration (i.e., 100\% confidence on all correct examples and 0\% confidence on all wrong examples), the averaged confidence scores shall be equal to the accuracy of the model on the data.

We can observe that, compared to the standard prompting, \emph{all other prompting methods suffer even more from over-confidence.}
Similar to well-observed \emph{anchoring bias} in human behavior~\cite {doi:10.1126/science.185.4157.1124, FURNHAM201135,lieder2018anchoring}, the generated thoughts may also mislead the model to be over-certain of its own answer. For example, the generated thoughts such as \textit{from the steps we can conclude that} can potentially make the model confident with its answers. If the output answer is wrong, the instance-level error from these high-confidence cases will be high. In the following section, we will investigate if treatment towards human \emph{anchoring bias} can be extended to model confidence calibration.

\section{Fact-and-Reflection Prompting}
\label{sec:far}

\subsection{Motivation and Design}

There have been studies in cognitive science on how to mitigate the over-confidence caused by anchoring-bias~\cite{BLOCK1991188, GEORGE2000195}, and the ideas have also been recently introduced to AI research in the context of decision-making ~\cite{10.1145/3491102.3517443}. 
One key insight is: \emph{Providing multiple facts, instead of conducting reasoning directly with the first acquired fact, can help reduce the bias.}
We take inspiration from this finding and propose to improve confidence calibration by \emph{decomposing the fact-acquiring and reflective reasoning steps during prompting}.
We expect this framework to encourage the inclusion of multi-perspective facts in the context before reasoning toward the answer. 
We denote it as \textbf{F}act-\textbf{a}nd-\textbf{R}eflection (\textbf{FaR}) prompting, which can be viewed as a new form of step decomposition prompting with constrained decomposition.


Specifically, in FaR prompt context, we guide the models with facts, sources of facts, and reflective reasoning conditioned on the facts before extracting confidence. In the CoT-style prompting, the answer $A$ is sampled from the $p (A | Q, T, \theta)$~\citep{Dohan2022LanguageMC}, where $Q$ denotes the query, $T$ denotes the thoughts generated by the model and $\theta$ denotes the model parameters. In FaR, we sequentially acquire two component thoughts $T_{\text{f}}$ and $ T_{\text{r}}$ (i.e., fact and reflection steps in Figure~\ref{fig:fsr}) regarding the known facts and reflective reasoning of the model about the questions. 
Motivated by~\citet{weller2023according}, we believe the trustworthiness of the sources can help stabilize the model reflection $T_{\text{r}}$. For model generated $T_{\text{f}}$, we add a sub-step to ask the model to elicit the known sources of the generated facts.

Upon acquiring the component thoughts, The final answer $A$ is sampled from $p (A | Q, T_{\text{f}}, T_{\text{r}}, \theta)$, i.e., the model generates the answer with the final step prompt including thoughts at each step. 

We design FaR prompting to be orthogonal to confidence extraction so that it can work together with different confidence extraction methods. The results of the step questions and the final answer will be utilized in the prompt to extract confidence.

In an ideal case, the model-generated ``facts'' should be verified by humans. However, human annotations are not always available and are hard to collect. To address this, we propose to prompt the models to generate relevant internal knowledge that may potentially help answer the questions as ``facts''. We conduct a pioneering study on introducing external verification in Section~\ref{sec:over_confidence}.

\subsection{Performance and Ablations}
\begin{table}[t]
\small
    \centering

    \begin{tabular}{l|c|c}
    \toprule
        Prompting Method  & ECE $\downarrow$ & MacroCE $\downarrow$\\
        \midrule
        Standard  & 30.3 & 54.6 \\ 
        Self-Ask (aggregate)  & 26.0 & 66.0 \\
        \midrule
        FaR (fact-only,no-source)  & 26.4 & 72.3 \\
        FaR (fact-only)  & 29.5 & 70.5 \\
        FaR (no-source)  & 28.4 & 60.5 \\
        FaR (+\textit{explain}) & 27.4 & 71.7 \\
        \midrule
        FaR (final)  & \textbf{22.8}  & \textbf{47.0}\\
         \bottomrule
    \end{tabular}
    \caption{Expected Calibration Error (ECE) and Macro-average Calibration Error (MacroCE) of different ablations of FaR. Standard and Self-Ask (aggregate) are provided for reference.   ``$\downarrow$'' denotes that lower is better. The best performance of each column is in \textbf{bold}.} 
    \label{tab:ablation}
\end{table}

We further study the generalized case with model-generated knowledge as $T_{\text{f}}$. 
We compare FaR and its different variants with the strongest baselines in Table~\ref{tab:prompt_style} (i.e., entries with the lowest calibration errors measured in ECE and in MacroCE), which are Self-Ask (aggregate) and Standard, respectively. Table~\ref{tab:ablation} shows that \textbf{FaR prompting brings improvements on both calibration metrics.}

We also include an extensive ablation study over the prompt components of FaR: \textit{fact-only} denotes the variant that we do not conduct the reflection step; \textit{no-source} is where we remove source generation (i.e., no ``what are the sources...'' step in Figure~\ref{fig:fsr}); \textit{+explain} changes the final prompt from ``Answer:'' to ``Explain and Answer:'', which is motivated by \textit{Knowledge+Explain}. 
From the table, we can observe that, although all variants achieve lower ECE than Standard Prompting, our final design (denoted as \emph{FaR (final)}) achieves the best performance among them. On the other hand, for MacroCE, most variants get lower performance than the standard prompting. Therefore, all the components in FaR are important in achieving good performance in both metrics. 


\begin{figure*}[t]
    \centering
    \includegraphics[clip,trim={0cm 0.3cm 0cm 0cm},width=0.9\linewidth]{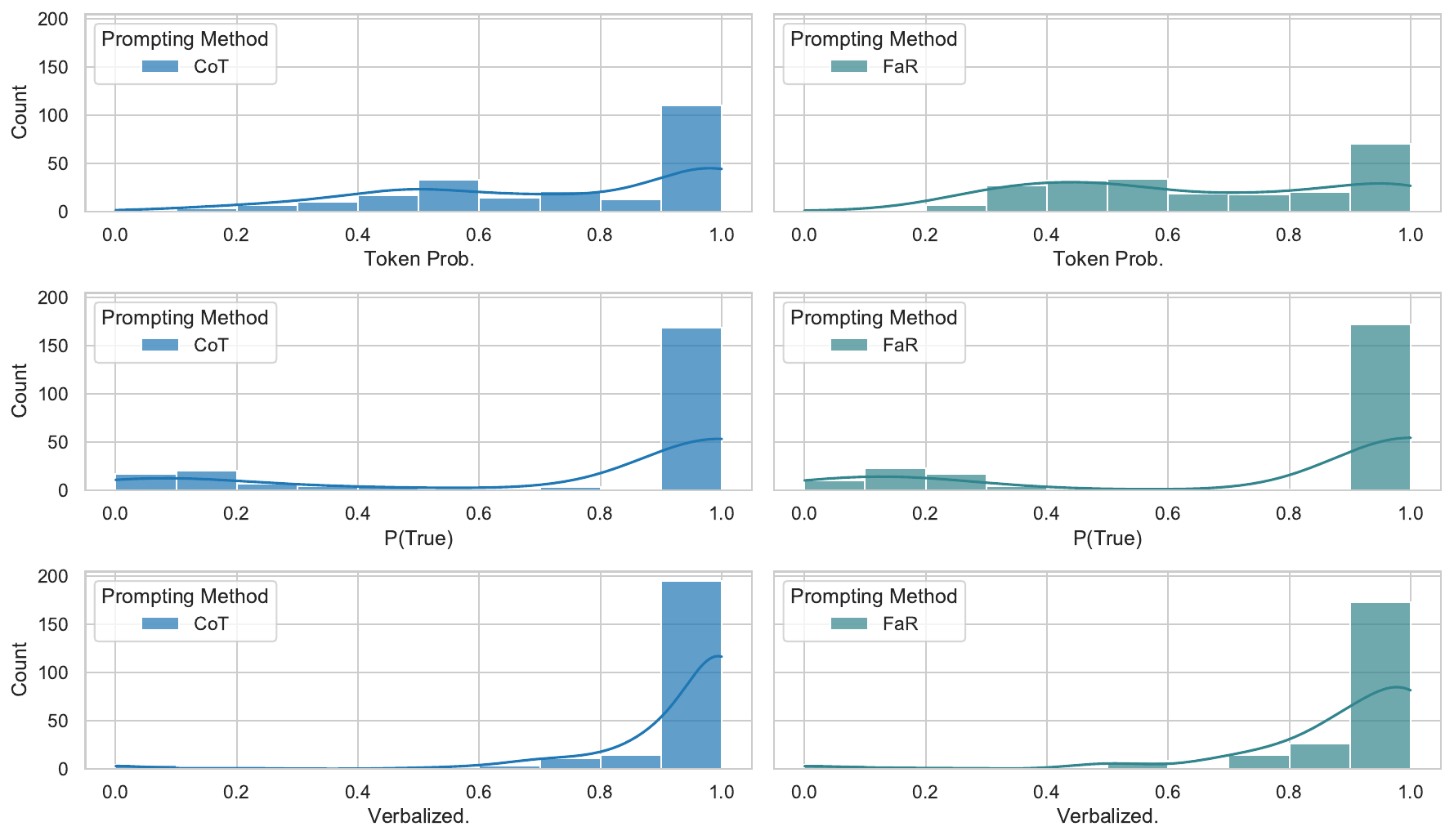}
    \caption{Distribution of the confidence scores (\textit{Token Prob., P(True), Verbalized}, in the top, middle, bottom, respectively), with different prompting styles, CoT (left) and FaR (right). Kernel Density Estimation (curved lines) is used to show smoothed distributions. FaR prompting helps lower the distribution mass on the high confidence side for different confidence extraction methods.}
    \label{fig:score_dist}
\end{figure*}

\subsection{What Impacted the Calibration?}
\label{sec:over_confidence}
The performance with ECE and MacroCE in Table~\ref{tab:prompt_style} and~\ref{tab:ablation} together presents that FaR \emph{does} mitigate the over-confidence issue analyzed in Section~\ref{sec:prompt_methods}, with both the correct and wrong predictions considered in the design of MacroCE.

We now analyze \emph{how} FaR prompting shifts the confidence distribution of the LLM generation. We compare it with CoT, as it is one of the best-performing models in the Step Decomposition category (Table~\ref{tab:prompt_style}). In addition, FaR can be regarded as a specially structured step decomposition prompting that explicitly disentangles the steps of knowledge self-prompting and reflective reasoning. 

Figure~\ref{fig:score_dist} presents the specific distribution of confidence scores extracted by  Token Prob., P(True), and Verbalized, respectively, smoothed by Kernel Density Estimation.
We can observe that FaR \textbf{mitigates the overconfidence issue} that we aspire to resolve for different confidence extraction methods --- it damps down the peak values and moves the density mass to the left side (lower confidence). That said, it still suffers from overconfidence. 
Such behavior partially explains the better confidence calibration results we observed in Section~\ref{sec:prompt_methods}. 
Even the density mass of the verbalized confidence is much balanced with FaR prompting, which greatly enhances the usability of such a method for black-box language models that have no access to the output token probabilities.
Again, consistent with our findings in Section~\ref{sec:prompt_methods},
various confidence extraction methods still suffer from \textit{over-confidence} issue, i.e., the confidence scores are higher than the target model performance, especially Verbalized Confidence. With Verbalized Confidence, even though FaR helps reduce over-confidence, further effort shall be made in future work to make it better calibrated.


\begin{table}[t]
\small
    \centering
    \begin{tabular}{p{3.2cm}|p{3.8cm}}
    \toprule
        Question (Label) & Model Output\\
         \midrule
         Would Persephone be a good consultant to a landscape architect? (\Green{True}) & False. There will need to be further research.\\
         \midrule
         Would an owl monkey enjoy a strawberry? (\Green{True}) & It is not possible to answer with current evidence this question.\\ %
         \midrule
         Does Post Malone have a fear of needles? (\Red{False}) & False, but there is not yet sufficient evidence to answer. \\%
         \midrule
         Should a Celiac sufferer avoid spaghetti? (\Green{True}) & False (It depends on the ingredients of the spaghetti) \\
         \bottomrule
    \end{tabular}
    \caption{Examples of the model expressing concerns. In addition to the output answer, the model also specifies whether the knowledge is sufficient or whether further conditions are needed to make the prediction. See Table~\ref{tab:full_far_example} for detailed outputs for other steps, e.g., diverse model-generated sources.}
    \label{tab:express_concern_example}
\end{table}


        


\paragraph{Expressing Concern.} How is FaR mitigating overconfidence? 
As shown in Table~\ref{tab:express_concern_example}, qualitatively, we observe an interesting phenomenon of the model outputs, named \textit{Expressing Concern}. With FaR prompting, besides outputting the answer, the model further specifies its thoughts on the answers, including (i) if the current evidence provided in the context is enough to answer the question; (ii) if a specific condition should be given to answer the question.

Comparing the probability of the model expressing concern using different prompts, 
FaR prompting can inspire the model to express concerns, compared to using the original CoT prompting (in 8.8\% vs 3.9\% examples). If we further remove the constraint on choosing one answer in the instruction, the model will elaborate further comments besides giving the answers (e.g., ``False. there is not yet sufficient evidence'') in 59.2\% cases with FaR prompting (denoted as FaR(free)).
These complex answers are harder for performance computation than simple short answers, but give a clear picture for reviewing the model comments on the given questions. For example, as shown in Table~\ref{tab:express_concern_example}, the model correctly answers that Post Malone does not have a fear of needles \footnote{The model supports its educated guess with Post Malone's numerous tattoos in the fact step. See Table~\ref{tab:full_far_example} for details of fact and reflection steps of this and other examples.}, but it still expresses concerns about insufficient evidence. Upon the concerns, users can choose to find further evidence not in the model training data to help verify the guess (e.g., some video interviews). Yet the definition of ``good'' concerns requires future investigation.

\begin{figure}[!t]
    \centering
    \includegraphics[clip,trim={0cm 1cm 0cm 0cm},width=0.95\linewidth]{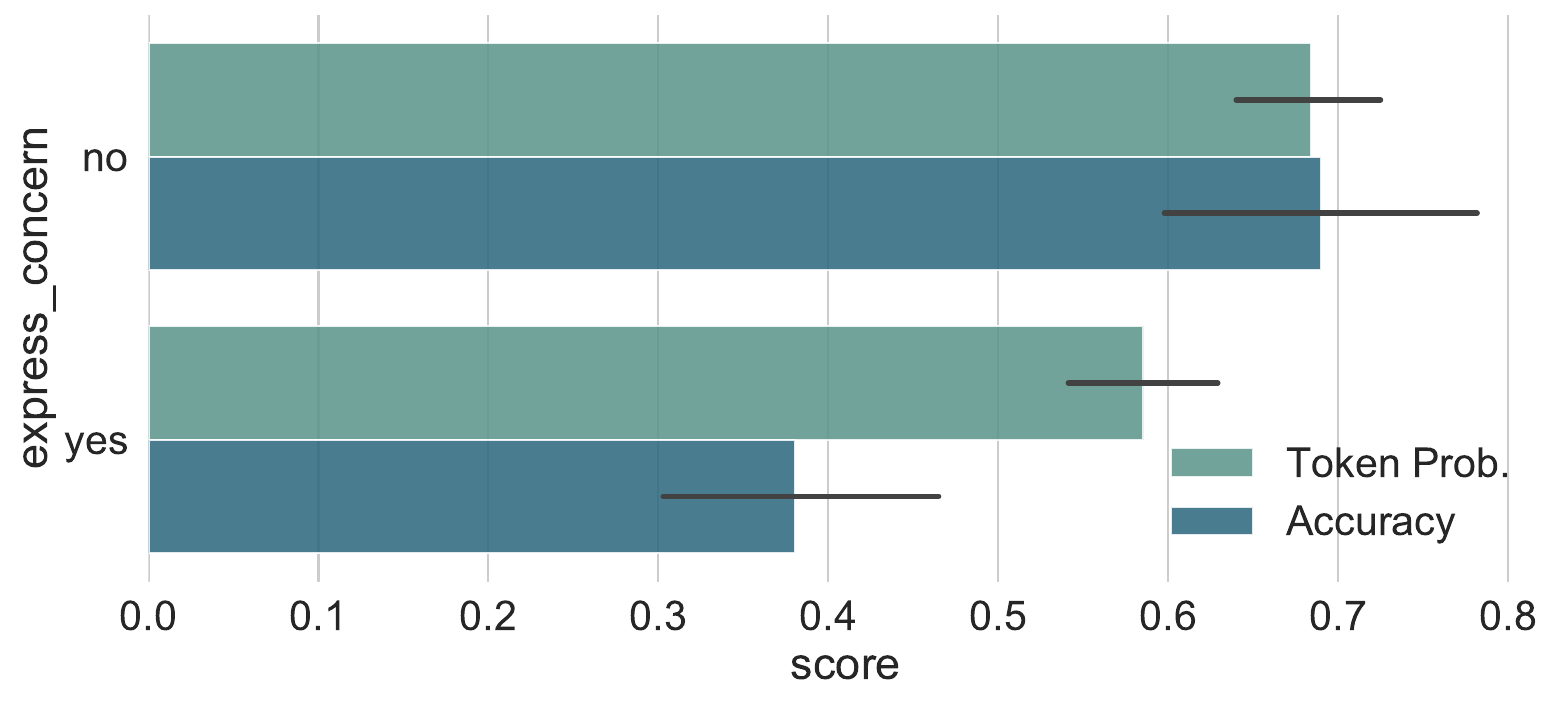}
    \caption{The confidence (extracted via token probability) and the accuracy when the model expresses (Lower) and does not express (Upper) concerns. With FaR (free) prompting, LLMs are more inclined to express concerns when the confidence and accuracy are low.}
    \label{fig:express_concern}
\end{figure}

Notably, as shown in Figure~\ref{fig:express_concern}, the expressing of concerns in the output typically co-occurs with lower confidence scores, which in turn co-occurs with lower accuracy on the questions. This implies that the model exhibits better confidence calibration --- with FaR prompting, the model tends to express concern on \textbf{hard examples}. 
As a result, it can further act as a kind of AI feedback~\cite{bai2022constitutional} for self-improvement by identifying difficult instances. For example, such a signal can trigger an iterative application of retrieval augmentation for checking and correcting the facts and sources provided by the models ~\cite{chen-etal-2021-improving}.

When extracting the statistics in Figure~\ref{fig:express_concern} from FaR(final) and CoT, on examples that the model expresses concerns, the accuracy is 25\% (FaR) vs. 60.2\% (CoT); on examples that models do not express concerns, the accuracy is 67.0\% (FaR) vs. 69.9\% (CoT). That is, FaR identifies examples with lower overall accuracy than CoT when concerns are expressed.


\paragraph{Addressing hard examples.} We further simulate the scenario of detecting hard examples that are not answerable with LLM's internal knowledge and then using external knowledge as augmentation. We only sample and apply retrieval augmentation to the examples where the model expresses concerns. 
The treatment is done by adding corresponding external knowledge to the hard examples in the context, with the same setting and external knowledge from~\citet{zhao2023thrust}\footnote{Further analysis on this setting incorporating external knowledge is in the appendix (Section~\ref{sec:verification_with_human}) .}. %

We compare this sampling strategy with randomly sampling the same portion of examples (i.e., control) and check the performance gain. With FaR(final) as the prompting method, we can observe a 68.0\% performance improvement in accuracy through external augmentation on hard examples that the model expresses concerns about. In contrast, randomly sampling the same portion of examples and applying augmentation only achieved a 15.0\% performance improvement in accuracy, implying that FaR(final) identifies the instances demanding the external knowledge more accurately by checking if the model \textit{expresses concerns}.


\begin{table}[t]
\small
    \centering
    \begin{tabular}{l|c|c}
    \toprule
        Prompting Method  & ECE $\downarrow$ & MacroCE $\downarrow$\\
        \midrule
        \multicolumn{3}{c}{ \textit{Vicuna-13b}}\\
        \midrule
        Standard  & 19.4 & 35.5 \\ %
        FaR  & \textbf{11.1} & \textbf{33.9} \\ %
        \midrule
        \multicolumn{3}{c}{ \textit{Baichuan-2-13b}}\\
        \midrule
        Standard  & 17.1 & 64.3 \\ %
        FaR  & \textbf{10.1} & \textbf{48.4} \\ %
        \midrule
        \multicolumn{3}{c}{ \textit{Llama-2-13b}}\\
        \midrule
        Standard  & 19.3 & 71.9 \\ %
        FaR  & \textbf{15.4} & \textbf{54.2} \\ %
         \bottomrule
    \end{tabular}
    \caption{Expected Calibration Error (ECE) and Macro-average Calibration Error (MacroCE) of different prompting methods and different backbone large language models. The down arrow implies the lower, the better. The best-performing entry on each column is marked in \textbf{bold}.} 
    \label{tab:model_ablation} 
\end{table}

\subsection{Generalizing to other language models}
\label{sec:generalize_backbone}
In this section, we generalize our experiments beyond GPT-3 to test the robustness of FaR across other language models. We follow the previous settings and conduct experiments on StrategyQA, with Token Prob. as the confidence extraction method.
We extend our experiments with Vicuna~\cite[\texttt{Vicuna-13b-v1.3},][]{zheng2023judging}, Baichuan~\cite[\texttt{Baichuan2-13B-Chat},][]{baichuan2023baichuan2}, LLama 2~\cite[\texttt{Llama-2-13b-chat-hf},][]{touvron2023llama}.
Our inference structure is built with vLLM~\cite{kwon2023efficient}. 

From Table~\ref{tab:model_ablation}, we can observe that, compared to Standard Prompting, FaR consistently leads to reduced calibration errors, measured by either ECE or MacroCE, which suggests the generalizability of FaR towards open-source language models.

\section{Related Work}\label{sec:related_work}

\paragraph{Prompting Large Language Models.} Recent research~\citep{brown2020language,kojima2023large} on large language models shows that in-context learning (ICL) achieves great effectiveness in using models as few-shot or zero-shot reasoners. Different styles of prompting such as Knowledge prompting~\citep{liu-etal-2022-generated}, Chain of Thought (CoT) prompting ~\citep{ weichain}, Self-Consistency prompting~\citep{wang2023selfconsistency}, Self-ask~\citep{press2023measuring}, Tree-of-Thought prompting~\citep{yao2023tree}, and Skill-in-Context~\citep{chen2023skills} are then proposed to guide the model to elicit its knowledge for reasoning in different ways. 

Most previous work mainly focuses on how such a prompting method influences the model performance on various tasks. In this paper, we compare how confidence calibration is influenced by different prompting methods.

\paragraph{Confidence Calibration of LLMs.} Extracting honest and constructive confidence scores of large language models is considered an important step towards building faithful and trustworthy AI systems~\citep{desai-durrett-2020-calibration,si2022prompting}. Many methods have been proposed recently to get reliable confidence scores with different suffix prompts added after outputting possible answers, such as a follow of True or False multiple choice question ~\cite{kadavath2022language}, asking models to describe the likelihood of the answer~\cite{lin2022teaching}, and describing the likelihood that the answer is correct with demonstrations~\cite{tian2023just}. 
However, it remains unclear how robust the methods are and how good they are comparatively. Our paper proposes FaR prompting as an orthogonal method to improve calibration and compare different extraction methods with our test bed.

Recently, \citet{yang2023alignment} discuss the honesty problem of models as part of the alignment. \citet{qian2023merge} study the confidence change when there is a conflict between in-context and model internal knowledge. Another line of work links model confidence with human confidence~\citep{zhou2023navigating, Steyvers2024TheCG, Zhou2024RelyingOT}. In our paper, we refer to the model trustworthiness based on confidence calibration.

\section{Conclusion and Discussion}
We closely examined how different prompting methods influence confidence calibration and found that over-confidence may lead to bad instance-level calibration. To address that, we propose FaR prompting, which decomposes the fact elicitation and reflective reasoning steps, and shows that it provides a good way to calibrate the model performance across different metrics.
Our further analysis reveals that the reasons behind the performance can be that FaR prompting elicits the model to generate more \textit{honest} answers, via \textit{expressing concerns} in lower confidence situations.

We encourage the designers of future prompting methods to evaluate their influence on the confidence calibration in addition to the performance. Meanwhile, we further suggest that future confidence extraction methods should take into account their robustness to different prompting methods.

\section*{Limitations}
\noindent\textbf{Extension to Human Instruction Datasets.} So far, we have conducted our experiments on question-answering datasets and shown the effectiveness of current prompting methods, confidence extraction methods, and our FaR prompting. Our method may be examined on the human instruction datasets as well, such as the Human Eval dataset ~\cite{selfinstruct}. However, since the targeted answers are in free form (e.g., ranging from designing a personal profile to writing a program for quick sort), it may require very different evaluation metrics beyond simple accuracy, which calls for further study and is thus beyond the scope of this paper. Multi-perspective evaluation of the output can be necessary, such as the annotations provided in \cite{malaviya23expertqa}. We consider such extension to be an important future work.

\noindent\textbf{Inner Model Dynamics.} The confidence extraction methods compared in this paper are mainly based on signals from the model's logits or final sampled tokens. It remains unclear how prompting methods will influence the model uncertainty estimated from inner model dynamics, e.g., the tree structure-ness of a sentence inside transformers~\cite{DBLP:conf/iclr/MurtySAM23}. We still study the LLM model as a black box and do not closely examine how the model's internal working mechanisms are influenced by the FaR prompting. One future direction is to study the confidence calibration problem by using mechanistic interpretation~\citep{olsson2022context}. We will leave such internal interpretability work as a future work.

\noindent\textbf{Model Elicited Sources.}
In this paper, we explore how model-generated knowledge sources help improve model confidence calibration.
In the Appendix, we present the detailed outputs for all steps of FaR, such as diverse model-generated sources that include but are not limited to URLs, in Table~\ref{tab:full_far_example}.  For example, in the last row of Table~\ref{tab:full_far_example}, the model elicits the sources as \textit{the Mayo Clinic and the National Institute of Diabetes and Digestive and Kidney Diseases (NIDDK)}. The model then autonomously comments on whether these sources are reputable, which plays the role of verbal confidence extraction on generated facts. 
However, \citet{gao2023enabling} directly generating sources based on facts may lead to inaccurate sources. Exploring the relation between the source's factualness and confidence calibration can be an important future direction.

\section*{Ethical Considerations}
Our datasets are written by professional annotators or extracted from Wikipedia for the purpose of scientific research on question-answering systems. However, we observe no model outputs that use extremely sensational language or inappropriate and aggressive language. The questions and outputs collected are all in English, which can limit the generalizability of the performance of our pipeline.

\section*{Acknowledgments}
This work is initiated during Xinran's internship at Tecent AI Lab, Bellevue. Xinran is supported by the ONR Award N000142312840. The authors thank Zhengxuan Wu, Xuanyu Zhou, Ruixin Hong, Vijay Viswanathan, Chenyang Yang, and Christina Ma for their insightful feedback and anonymous reviewers for helpful discussions and comments. \looseness=-1


\bibliography{metric}
\bibliographystyle{acl_natbib}

\clearpage

\appendix

\section{Appendix}

\begin{figure*}[!t]
    \centering
    \includegraphics[clip,trim={0cm 13cm 1cm 0cm},width=\linewidth]{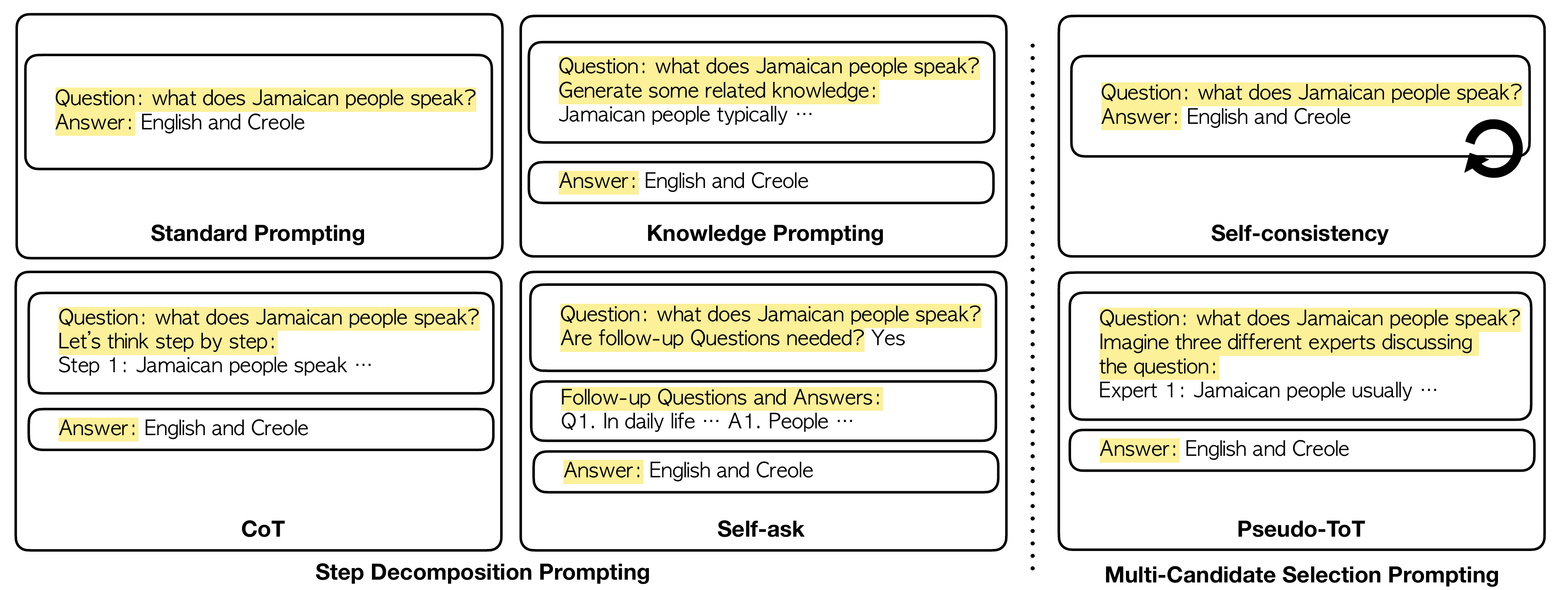}

    \caption{Different existing prompting methods to be analyzed. Each rounded rectangle represents one round of prompting. The intermediate answers will then be utilized in the final prompt to extract the answer to the original question. Highlighted text denotes the sentences provided to the model as the prompt. The loop icon in \textit{self-consistency} prompting denotes re-sampling with the same prompt multiple times.}
    \label{fig:prompt_examples}
\end{figure*}
\subsection{Examples of Baseline Prompting Methods}

In Section~\ref{sec:baselines}, we provide a detailed description of all the baseline prompting methods. Here we further illustrate their working mechanisms in Figure~\ref{fig:prompt_examples}. It shows how these methods are categorized into two major classes --- Step Decomposition and Multi-Candidate Selection --- based on the number of answer candidates are provided during the inference stage. All the baseline methods are evaluated on the original splits of StrategyQA (dev, 229 examples) and Web Questions (test, 100 examples), respectively. The data points we evaluated will be released upon acceptance.

\subsection{Details about Over-confidence or Under-confidence}

\begin{figure}[!h]
    \centering
    \includegraphics[width=\linewidth]{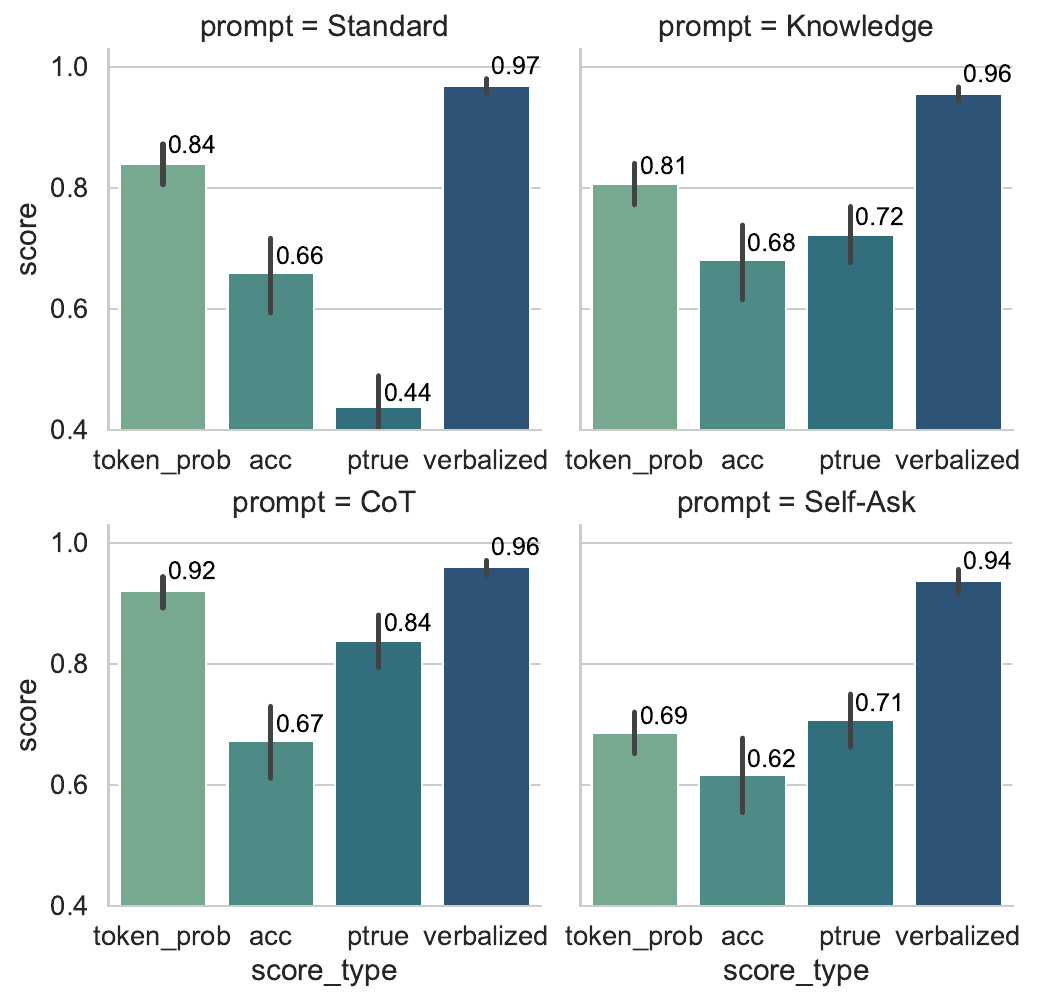}
    \caption{Accuracy and averaged confidence scores of different confidence extraction methods and prompting methods. In an ideal case with perfect calibration, the gap between these two scores should be zero.}
    \label{fig:over_under_confidence_details}
\end{figure}


As we discussed earlier, most of the prompting methods generally suffer from over-confidence. We now include some additional results to further show that this is true across different confidence extraction methods.

In Figure~\ref{fig:over_under_confidence_details}, we show the average confidence scores and the accuracies of different prompting methods, where the confidence score for each method is obtained by averaging the confidence scores of all samples. From Figure~\ref{fig:over_under_confidence_details}, we can observe that our conclusion on over-confidence exists generally across different confidence extraction methods and prompting methods. We can also identify that Verbalized Confidence suffers most from the over-confidence issue.

\subsection{Impact of Confidence Extraction Methods with Additional Metrics}
\label{sec:conf_extract_appendix}

\begin{table}[h]
\small
    \centering
    \begin{tabular}{p{1.8cm}|p{0.8cm}|p{0.9cm}|p{1.1cm}|p{1.1cm}}
    \toprule
        Conf. Extraction Methods & ECE-avg $\downarrow$ & ECE-wins $\uparrow$ & MacroCE-avg $\downarrow$ & MacroCE-wins $\uparrow$\\
         \midrule
         Token Prob. & 23.5 & 8 & 78.4 & 4 \\
         P(True) & 37.2 & 1 & 65.4 & 5 \\
         Verbalized & 40.6 & 0 & 101.9 & 0\\
         \bottomrule
    \end{tabular}
    \caption{Expected Calibration Error (ECE) and Macro-average Calibration Error (MacroCE) of different confidence extraction methods. Results are averaged over different prompting methods. Up (Down) arrows indicate the higher (lower) is the better.}
    \label{tab:conf_extraction_methods_appendix}
\end{table}

In Section~\ref{sec:conf_extract}, we compare different confidence extraction methods using the ECE and MacroCE (i.e., ECE-avg and MacroCE-avg) averaged over different prompting methods to measure the overall and instance-level performance (the lower, the better). To ensure that extreme values do not influence the conclusion, we further report the additional metric of ECE/MacroCE-wins in Table \ref{tab:conf_extraction_methods_appendix}. If a confidence extraction method achieves the lowest ECE/MacroCE-based errors with respect to a prompting method, it is marked as a win. All the scores are the average of 8 baseline methods in Table~\ref{tab:prompt_style} together with FaR (final).

From Table~\ref{tab:conf_extraction_methods_appendix}, we observe that Token Prob. achieves the best ECE for both the averaged scores and wins. In contrast, P(True) achieves the best MacroCE for both the averaged scores and wins, though the margin is not large. Therefore, the conclusion based on this new metric is consistent with the earlier findings.

\subsection{FaR with Human-Annotated External Knowledge}
\label{sec:verification_with_human}
\begin{table}[t]
\small
    \centering

    \begin{tabular}{l|c}
    \toprule
        Prompting Method  & ECE $\downarrow$ \\
        \midrule
        Standard  & 25.5\\ 
        FaR (human fact-only) & 14.8\\
        FaR (human fact + reflection)  & \textbf{13.6} \\
        
         \bottomrule
    \end{tabular}

    \caption{Expected Calibration Error (ECE) of using human-annotated facts in FaR, denoted by FaR (human). We use token probability as the confidence extraction method. The down arrow denotes that a lower score implies better calibration performance.} 
    \label{tab:motivation}
\end{table}

To further verify the idea that providing facts helps with confidence calibration, we incorporate multiple human-annotated facts (i.e., $T_f$) that are relevant to each question ($Q $) on the StrategyQA task~\citep{geva2021strategyqa}. The benchmark in this experiment is not the mix of StrategyQA and WebQ since human-annotated supporting facts are only provided in StrategyQA.
For example, for the question ``Did Aristotle use a laptop'', one piece of the fact can be ``The first laptop was invented in 1980''. Note that we assume the human-annotated facts are accurate, and therefore omit the step of generating the source. 
Incorporating human-annotated facts isolates the imperfection that may arise from model-generated facts and sources, which serves as a controlled ablation for examining the helpfulness of facts.
As shown in Table~\ref{tab:motivation}, including human-annotated facts in the context can improve the confidence calibration significantly. 

Furthermore, with the reflection step linking the facts and the final answer, the calibration can be further improved. However, human annotations are not always available and are hard to collect. To address this, our proposed FaR prompting elicits the models to generate relevant internal knowledge and sources that serve as the proxy for such golden human-annotated facts.

\subsection{Influence of Final-step Prompt Length}

\begin{figure}[!t]
    \centering
    \includegraphics[width=\linewidth]{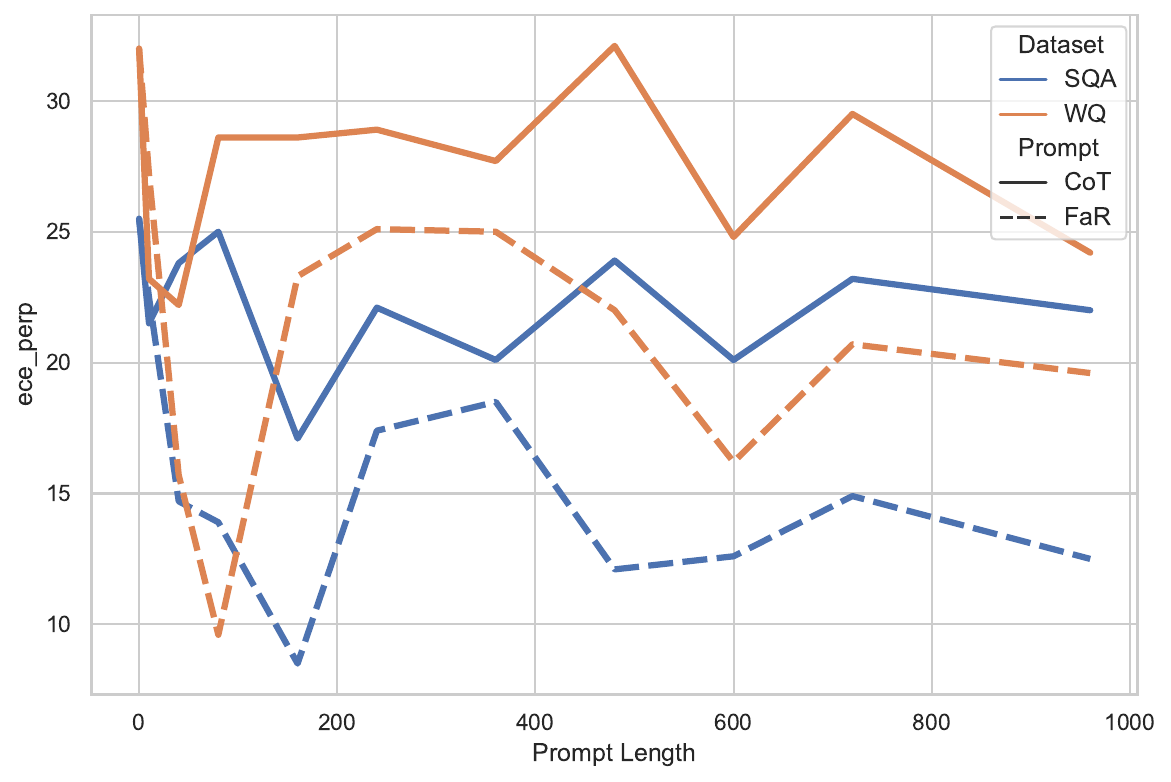}
    \caption{Expected Calibration Error (ECE) of CoT and FaR prompting with varying final prompt lengths on StrategyQA (SQA) and WQ.}
    \label{fig:prompt_len}
\end{figure}

We identify the final prompt length as another confounder in our experiments: although different prompting styles are initialized with the same max length to output the thoughts, the thoughts injected in the final prompt asking for answers to the original questions may still have varied length.

To remove the potential influence of actual prompt length, we further compare CoT and FaR with varying prompt lengths of the final step, asking the model for the answers. Figure~\ref{fig:prompt_len} presents the change of the ECE with respect to varying final-step prompt length. The table shows that FaR (dash lines) consistently outperforms CoT (solid lines) on different datasets and lengths. We can also observe that longer prompt length does not always lead to lower error, further validating our experimental results in Section~\ref{sec:prompt_methods} by removing the potential confounder: final prompt length.

\begin{table*}[t]
\small
    \centering


    \begin{tabular}{l|c|c|c|c|c}
    \toprule
        Method  & Accuracy & ECE-Token. & ECE-P(True)  & MacroCE-Token. & MacroCE-P(True) \\
        \midrule
        Standard & 60.3 & 25.5 & 35.1  & 67.8 & 41.5 \\
        \midrule
        Knowledge & 69.9 & 27.8 & 38.1  & 78.9 & 68.9\\
        Knowledge+Explain & 65.1 & 18.9 & 35.4  & \textbf{52.2} & 76.8 \\
        CoT & 67.2 & 25.0 & 34.3  & 82.6 & 42.0 \\
        Self-Ask & 60.7 & 17.6 & 35.3  & 58.5 & 74.7 \\
        Self-Ask (aggregate) & 60.4 & 17.0 & 35.0  & 60.0 & 72.0 \\
        \midrule
        Self-Con. & 63.3 & 35.9 & 33.5  & 97.9 & \textbf{37.2}\\
        Pseudo-ToT & 58.5 & 23.4 & 42.7  & 68.0 & 78.9 \\
        \midrule
        FaR(final) & 64.0 & \textbf{13.9} & \textbf{31.6}  & 52.9 & 41.0 \\
         \bottomrule
    \end{tabular}

    \caption{Expected Calibration Error (ECE) and Macro-average Calibration Error (MacroCE) of different prompting methods. The down arrow denotes the lower, the better. The best performance of each column is in \textbf{bold}. Token. and P(True) denote the use of Token Prob. and P(True) as the confidence extraction methods, respectively.} 
    \label{tab:full_performance} 
\end{table*}

\begin{figure}[t]
    \centering
    \includegraphics[width=\linewidth]{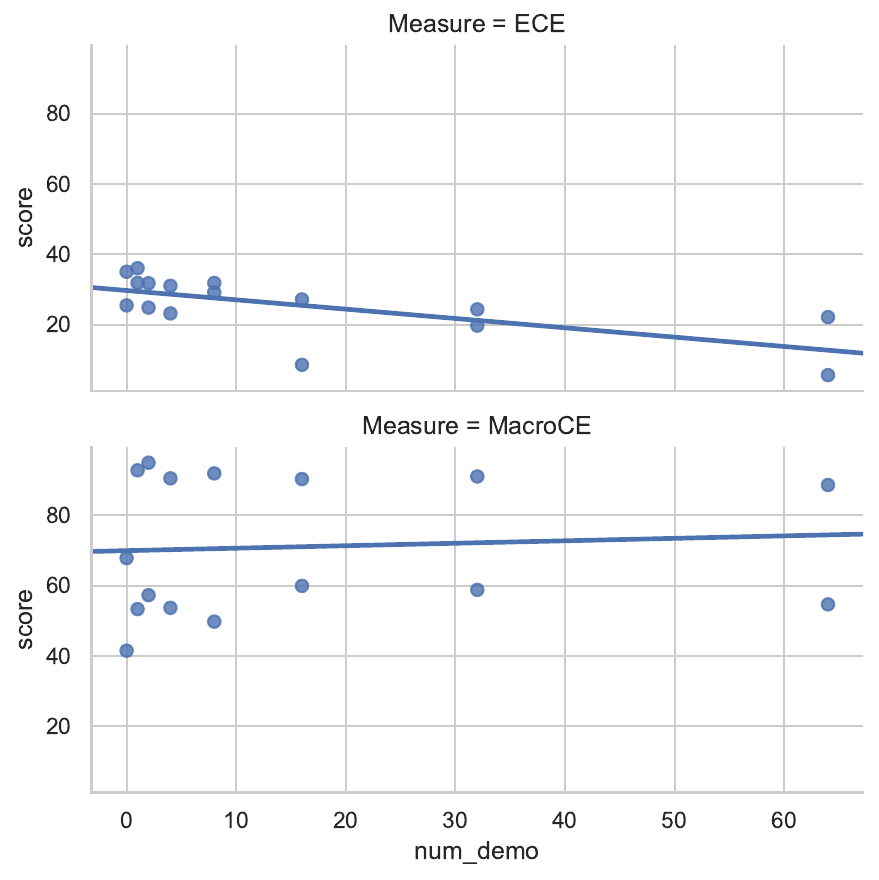}
    \caption{Expected Calibration Error (ECE) and Macro-average Calibration Error (MacroCE) of standard prompting with a varying number of demonstrations with Standard Prompting.}
    \label{fig:num_demo}
\end{figure}

\subsection{Influence of the Number of Demonstrations}

We identify the number of demonstrations in the prompt context as another important factor influencing the model calibration. 

Figure~\ref{fig:num_demo} presents the change of the ECE and MacroCE with respect to the changing number of demonstrations with standard prompting. From the figure we can observe that: increasing the number of demonstrations helps reduce the error measured by ECE but not the error measured by MacroCE.

The reason can be that similar to the influence of the number of demonstrations on performance, demonstrations help the models do well on the questions that are answerable and show good consistency and robustness on these questions. The overall calibration is also improved with more robust answers. However, for extremely hard cases where the model can not solve at all, increasing the number of demonstrations can not help solve the task, but may still lead to increased confidence, hence the instance-level calibration is not improved. 

\subsection{Full Results of ECE and MacroCE of different prompting methods}

In the main paper, we analyze the influence of prompting methods on confidence calibration from different perspectives, where for simplicity the results are presented by aggregating over different dimensions (e.g., averaging the metrics over different confidence extraction methods, or over different baselines). For completeness and for future reference, we report the full results of different prompting strategies (including our FaR) under different confidence extraction methods in Table~\ref{tab:full_performance}, where we report both ECE and MacroCE.

In terms of confidence calibration, Table~\ref{tab:full_performance} shows that FaR (final) performs the best or the second best in all settings. 
On the other hand, in terms of model performance in accuracy, we also observe that FaR (final) outperforms 5 out of 8 baselines, but is slightly worse when compared to CoT or Knowledge-based prompting. The reason can be that (i) further constraints are applied to the model-generated thoughts and limit the degree of freedom in search; (ii) the model tends to provide conservative answers. \textbf{The slight loss in accuracy may be viewed as a kind of ``calibration tax'' for achieving better calibration.} Nevertheless, as discussed in Section~\ref{sec:over_confidence}, the conservative answers with expressed concerns can trigger and guide external knowledge augmentation on the (hard) examples with significant performance gain.

\subsection{Full Examples of FaR Prompting}

\begin{table*}[h]
\small
    \centering
    \begin{tabular}{p{2.4cm}|p{2.8cm}|p{2.8cm}|p{2.8cm}|p{2.5cm}}
    \toprule
        Question (Label) & Facts & Sources & Reflection & Output\\
         \midrule
         Would Persephone be a good consultant to a landscape architect? (\Green{True}) 
         &1. Goddess of Spring and Vegetation: Persephone is often...
         2. Queen of the underworld... 
         & 1. Homer's Hymn to Demeter: this ancient text...
         2.Ovid's Metamorphoses: this collection of narratives...
         & 1. Connection to Vegetation and Renewal: Persephone's role... 2. Cultural and Historical Influence:...
         & False. There will need to be further research.\\
         \midrule
         
         Would an owl monkey enjoy a strawberry? (\Green{True}) 
         & 1. Diet of Owl Monkeys: Owal monkeys primarily have a diet... 2. Taste Preferences of Primates...
        & 1. Scientific journals and articles on primatology... (pubmed. ncbi.nlm.nih.gov) \quad\quad
        2. Zoological Association websites...
         &1. Sweet Taste Preference in Primates...\quad\quad
         2. Digestive Capabilities...
         & It is not possible to answer with current evidence this question.\\ %
         \midrule
         
         Does Post Malone have a fear of needles? (\Red{False}) 
         & 1. Tattoo Collection: he has a significant number of tattoos... 2. Public Statements:...
         & 1. Interviews and Profiles: publications like Rolling Stone, Billboard, GQ...
         2. Social Media: his accounts on platforms...
         & 1. Nature of Tattoos: tattoos are created by repeatedly inserting a needle... 2. Lack of Public Statements regarding a fear of needles:...
         & False, but there is not yet sufficient evidence to answer. \\%
         \midrule
         Should a Celiac sufferer avoid spaghetti? (\Green{True}) 
         & 1. Definition of Celiac Disease: Celiac disease is an autoimmune... 2. Gluten in Wheat and Gluten-free Alternatives:...
         & The recommendation on avoiding traditional spaghetti is supported by reputable sources like Mayo Clinic and NIDDK...
         & 1. Nature of Celiac Disease:... Health Implications of Gluten Consumption: continuous consumption of gluten...
         & False (It depends on the ingredients of the spaghetti) \\
         \bottomrule
    \end{tabular}
    \caption{Examples of the model output on the fact, reflection, and answer steps of FaR prompting. Detailed explanations on each point from the model are omitted for the presentation purpose.}
    \label{tab:full_far_example}
\end{table*}

In Figure~\ref{fig:fsr} and Table~\ref{tab:express_concern_example}, we demonstrate the general ideas about how FaR prompting works and how the model answers questions while expressing necessary concerns. Recall that FaR sequentially prompts the LLM for (i) facts, (ii) sources; (iii) reflection, and (iv) answers. In Table~\ref{tab:full_far_example}, we further provide the outputs at each of these steps for the examples in Table \ref{tab:express_concern_example}. We can observe that the output concern is relevant to the facts elicited in the context. For example, for the question, \textit{Should a Celiac sufferer avoid spaghetti?}, the model says \textit{it depends on the ingredients}, which relates to \textit{Gluten-free alternatives for spaghetti} that is discussed in the fact step output of the model.



    

\end{document}